\documentclass[journal,9pt]{IEEEtran}

\usepackage{amsmath,amssymb,enumerate,subfigure,multirow,hhline,color}
\usepackage{xcolor}
\usepackage{graphicx}
\usepackage{graphics}
\usepackage{amsthm}
\usepackage{algorithm}
\usepackage{algpseudocode}
\usepackage{tcolorbox}
\usepackage{epsfig,psfrag}
\usepackage{tabularx}
\usepackage{tabulary}

\usepackage[sort]{cite}

\usepackage{hyperref}
\hypersetup{breaklinks=true,colorlinks=true,linkcolor=blue,citecolor=blue}

\usepackage[noabbrev,capitalise,nameinlink]{cleveref}

\DeclareMathOperator{\argmin}{arg\,min}
\DeclareMathOperator{\argmax}{arg\,max}

\allowdisplaybreaks

\title{Simulation Studies on Deep Reinforcement Learning for Building Control with Human Interaction}

\author{Donghwan Lee, Niao He, Seungjae Lee, Panagiota Karava, and Jianghai Hu
\thanks{This material is based upon work supported by the National Science Foundation under Grant No.~1539527. Preliminary results of this paper have been published in~\cite{donghwan2018simulation,donghwan2018approximate}.}
\thanks{D. Lee is with the Department of Electrical Engineering, KAIST, Daejeon, 34141, South Korea {\tt\small
donghwan@kaist.ac.kr}.}
\thanks{N. He is with the Department of Computer Science,
ETH Z\"{u}rich, CH-8092 Z\"{u}rich, Switzerland {\tt\small
niao.he@inf.ethz.ch}.}
\thanks{S. Lee and P. Karava are with the Department of Civil Engineering, Purdue University, Purdue
University, West Lafayette, IN 47906, USA {\tt\small
lee1904@purdue.edu, pkarava@purdue.edu}.}
\thanks{J. Hu is with the Department of Electrical and Computer Engineering,
Purdue University, West Lafayette, IN 47906, USA {\tt\small
jianghai@purdue.edu}.} }
\begin{document}
\maketitle

\begin{abstract}
The building sector consumes the largest energy in the world, and
there have been considerable research interests in energy
consumption and comfort management of buildings. Inspired by
recent advances in reinforcement learning (RL), this
paper aims at assessing the potential of RL in building climate
control problems with occupant interaction. We apply a recent RL
approach, called DDPG (deep deterministic policy gradient), for
the continuous building control tasks and assess its performance
with simulation studies in terms of its ability to handle (a) the
partial state observability due to sensor limitations; (b) complex
stochastic system with high-dimensional state-spaces, which are jointly continuous and discrete; (c)
uncertainties due to ambient weather conditions, occupant's
behavior, and comfort feelings. Especially, the partial observability and uncertainty due to the occupant interaction significantly complicate the control problem. Through simulation studies, the
policy learned by DDPG demonstrates reasonable performance and
computational tractability.
\end{abstract}

\section{Introduction}

The building sector is known to consume around 41\%
of the energy in the United States~\cite{DOE2011} on an annual
basis. For this reason, the building climate control problem has
attracted much
attention~\cite{ma2012predictive,dounis2009advanced,ma2012fast}
over the past years. Its main goal is to balance
between the energy consumption and occupants' comfort in work
environments. In building spaces with occupants, uncertainties are
introduced by the action of occupants and incomplete knowledge of
their comfort feelings, which are significant in the thermal
dynamics of building
spaces~\cite{aswani2012reducing,oldewurtel2013importance,dobbs2014model,page2008generalised}
as well as in deciding the optimal control performance criteria.
An important issue that emerged recently is encoding
such information into the design process. However,
the building system with occupants is a complex stochastic system
with high-dimensional jointly continuous and discrete
state-spaces, and its real-world control tasks suffer from the
partial state observability due to sensor limitations. Therefore,
establishing a scalable control design framework is of prime
importance. Inspired by recent advances in reinforcement learning
(RL)~\cite{mnih2015human,silver2016mastering}, the main goal of
this paper is to assess the potential of RL in building climate
control problems with occupant interaction.

Optimal control designs for stochastic
systems have been a fundamental research field with a long
tradition, for instance, the linear quadratic Gaussian
problem~\cite{bertsekas1996neuro}, convex optimization-based
design for Markovian disturbances~\cite{costa1999constrained},
stochastic model predictive control (SMPC) for
Gaussian~\cite{primbs2009stochastic} and
Markovian~\cite{patrinos2014stochastic,di2014stochastic}
disturbance/uncertainties, approximate dynamic programming
(ADP)~\cite{kolmanovsky2008discrete,johannesson2007assessing} for
hybrid electric vehicle powertrain management problems, and
scenario-based (or sample-based) approximation approaches for the
vehicle path-planning~\cite{blackmore2007robust,blackmore2010probabilistic},
aircraft conflict detection~\cite{prandini2000probabilistic}, and
stochastic MPC~\cite{schildbach2014scenario,calafiore2013robust}
to name a few. Most approaches mentioned above
formulate the design problems into optimization problems, which
however, cannot be easily generalized and scaled to address more
generic and complex tasks with high dimensions. On the other hand,
reinforcement learning (RL)~\cite{sutton1998reinforcement}
algorithms provide a sound theoretical/practical framework to
address the problem of how an agent learns to take actions to
maximize cumulative rewards by interacting with the unknown
environment. Recently, significant progress has been made to
combine the advances in deep neural network
learning~\cite{lecun2015deep} with RL~\cite{mnih2015human} to
approximate value functions in high-dimensional control- and
state-spaces, which has demonstrated successful performance on
various challenging control tasks, for example, Atari
games~\cite{mnih2015human}, AlphaGo~\cite{silver2016mastering},
and a variety of continuous control
problems~\cite{heess2015memory,silver2014deterministic,lillicrap2015continuous}.

For the building control, various stochastic control approaches
have been applied. A nonlinear optimization method was studied for
SMPC in~\cite{ma2012fast-b}, which was extended
in~\cite{ma2012fast} to deal with chance constraints. SMPC was
also applied in~\cite{oldewurtel2012use} with weather predictions,
and comprehensive guides and surveys were provided
in~\cite{ma2012predictive}. To cope with generic non-Gaussian
stochastic disturbances, a scenario-based SMPC was investigated
in~\cite{zhang2013scenario}. On the other hand, RL has been
studied in~\cite{henze2003evaluation,yang2015reinforcement,hafner2011reinforcement,liu2006experimental,yu2010online,cheng2016satisfaction}
to find a balance among energy savings, high comfort, and indoor
air quality. In particular, experimental and comparative
performance analysis of Q-learning was given
in~\cite{liu2006experimental,henze2003evaluation}. Tabular
Q-learning and neural network-based batch Q-learning
were implemented
in~\cite{yang2015reinforcement,hafner2011reinforcement}. Other
approaches include the fuzzy logic-based
Q-learning~\cite{yu2010online} and Q-learning approach to the
lightning and blind control with a user feedback
interface~\cite{cheng2016satisfaction}. Recently, RL algorithms
based on the deep Q-learning~\cite{mnih2015human} was applied to
the building problems in~\cite{wei2017deep,nagy2018deep}.

While RL approaches in the building literature are not entirely new, to the authors'
knowledge, the potential of recent
advances~\cite{heess2015memory,silver2014deterministic,lillicrap2015continuous}
has not been fully investigated so far. In particular, the
majority of RL-based
approaches~\cite{henze2003evaluation,yang2015reinforcement,hafner2011reinforcement,liu2006experimental,yu2010online,cheng2016satisfaction,wei2017deep,nagy2018deep}
in the building literature employ Q-learning algorithms, which is
suitable only with discrete action spaces. For Q-learning
algorithms with continuous action spaces, finding the greedy
policy requires an optimization of the value function over the
continuous action space at every time
step~\cite{lillicrap2015continuous}, which is computationally
infeasible in general. However, building spaces conditioned by a
VAV (variable air volume) system has continuous action spaces.

Recently, a policy gradient RL for continuous control tasks,
called DDPG (deep deterministic policy gradient)
algorithm~\cite{silver2014deterministic,lillicrap2015continuous},
was developed to overcome this challenge based on the actor-critic
algorithm~\cite{konda2000actor}. It uses deep neural networks for
both actor (deterministic policy) and critic (value function)
function approximations and apply a stochastic gradient descent
method to maximize the value function and minimize the Bellman
loss function. The focus of this paper is to assess the
performance of DDGP with simulation studies taking into account
occupant-building interactions. We incorporate the occupant
comfort feelings into the design and simulation studies.
Contrasting with existing approaches, a probabilistic occupant
thermal preference model~\cite{lee2017bayesian,lee2019inference} is used directly
for training the policy to maintain more comfort indoor air
temperature.

Finally, most real-life problems are partially observed, and the
building system is no exception. Most SMPC algorithms for the
building problems assume that the full state information is
available. However, in many building systems, the access to the
full system information, such as the ambient weather conditions,
is not possible due to limited sensors and building
infrastructures.  An additional contribution is the consideration
of the partial observability in building systems. DDPG algorithm
is applied with a direct observation, and its performance is
evaluated.

\section{Problem Formulation}\label{section:preliminaries}

\subsection{Markov decision process (MDP)}
In a Markov decision process with the state-space ${\cal S}:=\{ 1,2,\ldots ,|{\cal S}|\}$ and action-space ${\cal A}:= \{1,2,\ldots,|{\cal A}|\}$, the decision maker selects an action $a \in {\cal A}$ with the current state $s$, then the state transits to $s'$ with probability $P(s'|s,a)$, and the transition incurs a random reward $r(s,a,s')$, $P(s'|s,a)$ is the state transition probability from the current state $s\in {\cal S}$ to the next state $s' \in {\cal S}$ under action $a \in {\cal A}$, and $r(s,a,s')$ is the reward function. The decision maker or agent sequentially takes actions to maximize cumulative discounted rewards. A deterministic policy, $\pi :{\cal S} \to {\cal A}$, maps a state $s \in {\cal S}$ to an action $\pi(s)\in {\cal A}$. The Markov decision problem (MDP) is to find a deterministic or stochastic optimal policy, $\pi^*$, such that the cumulative discounted rewards over infinite time horizons is maximized, i.e.,
\begin{align*}
\pi^*:= \argmax_{\pi\in \Theta} {\mathbb E}\left[ \left.\sum_{k=0}^\infty {\alpha^k r(s_k,a_k,s_{k+1})}\right|\pi\right],
\end{align*}
where $\gamma \in [0,1)$ is the discount factor, $\Theta$ is the set of all admissible deterministic policies, $(s_0,a_0,s_1,a_1,\ldots)$ is a state-action trajectory generated by the Markov chain under policy $\pi$, and ${\mathbb E}[\cdot|\cdot,\pi]$ is an expectation conditioned on the policy $\pi$. The Q-function under policy $\pi$ is defined as
\begin{align*}
Q^{\pi}(s,a)=&{\mathbb E}\left[ \left. \sum_{k=0}^\infty {\alpha^k r(s_k,a_k,s_{k+1})} \right|s_0=s,a_0=a,\pi \right],\\
s\in& {\cal S},a\in {\cal A},
\end{align*}
and the corresponding optimal Q-function is defined as $Q^*(s,a)=Q^{\pi^*}(s,a)$ for all $s\in {\cal S},a\in {\cal A}$. Once $Q^*$ is known, then an optimal policy can be retrieved by $\pi^*(s)=\argmax_{a\in {\cal A}}Q^*(s,a)$. When the model is known, the problem can be solved by using the dynamic programming (DP), which tries to solve the Bellman equation~\cite{bertsekas1996neuro}. Otherwise, we can use model-free algorithms such as reinforcement learning (RL)~\cite{sutton1998reinforcement}, which is a class of model-free learning algorithms to find an optimal control of unknown systems by interacting with the unknown environment. For simplicity of the presentation, we only consider MDPs with discrete state and action spaces. However, the arguments in this paper can be generalized to MDPs with continuous state and action spaces or jointly continuous and discrete spaces with some modifications.

\subsection{Partially Observed MDP (POMDP)}
In real world applications, the full-state information is rarely available. The MDP is partially observed when the agent is unable to observe the state $s_k$
directly, and instead receives observations from the set ${\cal O}$ which
are conditioned on the underlying state $s_k$~\cite{heess2015memory}, i.e., there exists a probability distribution $P_O(o|s)$ of $o_k$ conditioned on the state. The sequential decision problem with this additional constraint is called the partially observable Markov decision problem (POMDP). In a POMDP with the state-space ${\cal S}:=\{ 1,2,\ldots ,|{\cal S}|\}$, action-space ${\cal A}:= \{1,2,\ldots,|{\cal A}|\}$, and observation space ${\cal O}:= \{1,2,\ldots,|{\cal O}|\}$, the decision maker selects an action $a \in {\cal A}$ with the current observation $o \sim P_O(\cdot|s)$ conditional on the current state $s$, then the state transits to $s'$ with probability $P(s'|s,a)$, and the transition incurs a random reward $r(o,a,o')$. Usually, the observation $o_k$ at time $k$ does not have the Markov property~\cite[pp.~63]{resnick2013adventures}, i.e., its
transition depends on all the past history of observations and
actions. For this reason, an optimal policy may, in principle, require access to the entire history~\cite{heess2015memory} and may in general be
non-stationary~\cite[Fact~4]{singh1994learning}. In this paper, we restrict
our attention to the deterministic stationary policy $\mu: {\cal O} \to {\cal A}$
mapping from $\cal O$ to the control space $\cal A$. To differentiate it from
the usual policy based on states, it will be called an output-feedback policy.
The decision problem under the output-feedback policy is to find a suboptimal policy, $\mu^*$, such that the cumulative discounted rewards over infinite time horizons is maximized, i.e.,
\begin{align*}
\mu ^* : = \argmax _{\pi  \in \Pi } {\mathbb E}\left[ {\left. {\sum_{k = 0}^\infty  {\alpha ^k r(o_k ,a_k ,o_{k + 1} )} } \right|\mu } \right],
\end{align*}
where $\Pi$ is the set of all admissible control policies $\mu:{\cal O}\to {\cal A}$.

In general, RL and DP approaches are applicable only to MDPs with fully observable states because the DP convergence is based on the Markov property, which is lacking in POMDPs. Nevertheless, a naive application of RLs to POMDPs is known to demonstrate reasonable performances in practice~\cite{singh1994learning,heess2015memory}. In particular,~\cite{singh1994learning} proved that some RL algorithms solve a modified Bellman equation and compute an approximate policy. The building control problem under our consideration is an POMDP because of the limited access to the full system information, such as the ambient weather conditions. In the next sections, we briefly introduce a naive DDPG algorithm, which will be applied with a direct observation afterwards.

\section{Deep RL algorithms}
RLs can be interpreted as sample-based stochastic dynamic programming (DP) algorithms that solve the Bellman equation. For many important real world problems, the computational requirements of the DP and RL are overwhelming as the state and control spaces are very large. Many RL algorithms~\cite{sutton1998reinforcement} use
the parameterized Q-factor $Q(\cdot,\cdot|\theta^Q)$, and
approximately solve the Bellman equation with a stream of state observations when the state transition model is not known, as is often the case in real-world applications. In particular, Q-learning~\cite{watkins1989learning} solves the
Bellman equation with samples/observations
by minimizing the loss
\begin{align*}
&L(\theta^Q):={\mathbb E} [(Q(s_k,a_k|\theta^Q)- y_k)^2],
\end{align*}
where $y_k:=r(s_k,s_k,s_{k+1}) + \alpha \max_{a\in {\cal A}} Q(s_{k+1},a)|\theta^Q)$, and the expectation is taken
with respect to $s_{k+1},a_k,s_{k+1}$. Recent RL
approaches use large-scale deep neural networks~\cite{lecun2015deep}, for instance, the deep Q-learning~\cite{mnih2015human} and deep deterministic policy
gradient (DDPG)~\cite{silver2014deterministic,lillicrap2015continuous}, to
approximate value functions in high-dimensional spaces. The DDPG is an actor-critic algorithm~\cite{konda2000actor,sutton2000policy} with neural networks, where a parameterized deterministic control policy (actor),
$\pi(s|\theta^\pi)\in {\cal A}, s\in {\cal S}$, instead of the greedy policy and the parameterized Q-factor $Q^{\pi(\cdot|\theta^\pi)}(s,a)$ (critic) is learned. In particular, it simultaneously performs the two optimizations
\begin{align*}
\theta_*^Q:=&\argmin_{\theta^Q} L(\theta^Q)\\
:=&{\mathbb E}_{(s_k ,a_k,r_k) \sim U(R)} [(Q(s_k,a_k |\theta^Q)-y_k)^2],
\end{align*}
where $y_k:=r(s_k,a_k,s_{k+1})+\alpha Q(s_{k+1},\pi(s_{k+1}|\theta^\pi)|\theta^Q)$, and
\begin{align*}
&\theta_*^\pi:=\argmax_{\theta^\pi}{\mathbb E}_{s_k}[Q(s_k,\pi(s_k|\theta^\pi)|\theta^Q)].
\end{align*}

Both the actor and critic parameters are updated by using stochastic gradient descent steps. A naive application of the DDPG to POMDPs is summarized in~\cref{algo:DDPG}.
\begin{algorithm}[h]
\caption{DDPG algorithm~\cite{lillicrap2015continuous}}
\begin{algorithmic}[1]
\State Randomly initialize the critic $Q(\cdot,\cdot|\theta^Q)$
and actor $\mu(\cdot |\theta^\pi)$ networks with weights
$\theta^Q$ and $\theta^\pi$, respectively.

\State Initialize the so-called target critic network
$Q'(\cdot,\cdot|\theta^{Q'})$ and the target actor network
$\mu'(\cdot|\theta^{\mu'})$ with weights $\theta^{Q'} \leftarrow
\theta^Q,\theta^{\mu'}\leftarrow \theta^\mu$.

\State Initialize the replay buffer ${\cal R}$.

\For{Episodes from $1$ to $M$}

\State Initialize a random process ${\cal N}:= (e_k)_{k=0}^\infty $ for action exploration.

\State Receive the initial state $s_0$.

\For{$k \in \{0,1,\ldots,T - 1\}$}

\State Select an action $a_k= \pi(s_k|\theta^\pi)+e_k$ according to the current policy and exploration noise.

\State Execute the action $a_k$, receive the reward $r(s_k,a_k,s_{k+1})$ and observe $s_{k+1}$.

\State Store the transition $(s_k,a_k,r(s_k,a_k,s_{k+1}),s_{k+1})$ in ${\cal R}$.

\State Uniformly sample a random minibatch of $N$ transitions
$(s_i,a_i,r(s_i,a_i,s_{i+1}),s_{i+1})$
from ${\cal R}$.

\State Set $y_i= r(s_i,a_i,s_{i+1}) + \alpha Q'(s_{i+1},\pi'(s_{i+1}|\theta^{\pi'})|\theta^{Q'})$.

\State Update the critic by minimizing the loss $L: =
\frac{1}{N}\sum_{i=1}^N {(y_i-Q(s_i,a_i|\theta^Q))^2}$.

\State Update the actor policy using the sampled policy gradient
\begin{align*}
&\nabla_{\theta^\pi} J^{\pi(\cdot|\theta^\pi)} \\
& \cong \frac{1}{N}\sum_{i=1}^N {\left. \nabla_{\theta^\pi} \pi
(s_i|\theta^\pi)\nabla_a Q(s_i,a|\theta^Q) \right|_{a
= \pi(s_i|\theta^\pi)}}.
\end{align*}

\State Update the target networks
\begin{align*}
&\theta^{Q'} \leftarrow \tau \theta^Q+(1-\tau)\theta^{Q'},\quad
\theta^{\pi'}\leftarrow \tau \theta^\pi + (1-\tau)\theta^{\pi'},
\end{align*}
where $\tau \in (0,1)$ is the update step size.

\EndFor

\EndFor

\end{algorithmic}
\label{algo:DDPG}
\end{algorithm}

The algorithm can be applied to POMDPs with $(o_k,o_{k+1},\pi)$ replaced with $(s_k,s_{k+1},\mu)$.

\section{Building Control with Occupant Interaction}\label{section:building-control-problem}
\subsection{Building Model}
\begin{figure}[h]
\centering\epsfig{figure=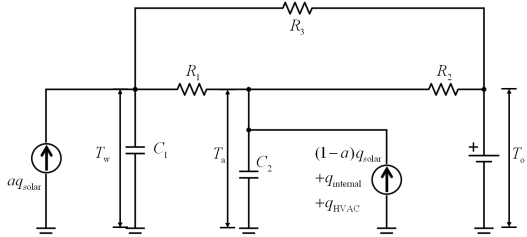,width=7cm} \caption{RC
circuit analogy}\label{figure:RLC-circuit-diagram}
\end{figure}

We consider a $3{\rm m} \times 3{\rm m}$ private office space with
a $2.5 {\rm m}^2$ south facing window, and its RC
(resistor-capacitor) circuit analogy is given
in~\cref{figure:RLC-circuit-diagram}. To reduce the order of the
model, we use one node for air in the room and another node
collecting all the thermal mass in the room,
where ${\bf T}_{\rm a}$ is the air temperature ($^{\circ} C$),
${\bf T}_{\rm o}$ is the outdoor air temperature ($^{\circ} C$),
${\bf T}_{\rm w}$ is the temperature of the aggregated mass node
($^{\circ} C$), ${\bf q}_{\rm solar}$ is the solar radiation
($W$), ${\bf q}_{\rm internal}$ is the internal heat ($W$), ${\bf
q}_{\rm HVAC}$ is the heating/cooling rate of the HVAC (heating,
ventilating, and air-conditioning) system ($W$). We assume that
the room is conditioned by a VAV (variable air volume) system so
that ${\bf q}_{\rm HVAC}$ directly affects ${\bf T}_{\rm a}$.
Since we use low order model, we assume that the air node includes
some portion of surfaces in the room which absorb radiative heat
and release the heat quickly to the air. To determine appropriate
values of the parameters of the circuit, we conducted a building
energy simulation with EnergyPlus 8.7.0 in~\cite{energyplus}, and
estimated the parameters minimizing the root-mean-square error
between the air temperatures calculated by the EnergyPlus
simulation and the low order model. The values of parameters are
summarized in~\cref{table:circuit-parameters}. Moreover, all
notations used to describe the building system are presented
in~\cref{table:buliding-notations}.
\begin{table}[h]
\caption{Values of the parameters of the circuit
in~\cref{figure:RLC-circuit-diagram}}
\begin{center}
\begin{tabular}{c c c}

\hline

Parameter & Value & Unit\\

\hline

$R_1$ & $0.0084197$ & $^\circ C/W$\\
$R_2$ & $0.044014$ & $^\circ C/W$\\
$R_3$ & $4.38$ & $^\circ C/W$\\
$C_1$ & $9861100$ & $J/^\circ C$\\
$C_2$ & $128560$ & $J/^\circ C$\\
$a$ & $0.55$ & --\\

\hline
\end{tabular}
\label{table:circuit-parameters}
\end{center}
\end{table}
\begin{table}[h]
\caption{Notations}
\begin{center}
\begin{tabular}{c c}

\hline

Notation & Meaning\\

\hline
${\bf T}_{\rm a}$ & Indoor air temperature ($^{\circ} C$)\\
${\bf T}_{\rm o}$ & Outdoor air temperature ($^{\circ} C$)\\
${\bf T}_{\rm w}$ & Temperature of the aggregated mass node
($^{\circ} C$)\\
${\bf q}_{\rm solar}$ & Solar radiation ($W$)\\
${\bf q}_{\rm internal}$ & Internal heat ($W$)\\
${\bf q}_{\rm HVAC}$  & Heating/cooling rate of the HVAC
system ($W$)\\
$\Delta t$  & Sampling time (min)\\

\hline
\end{tabular}
\label{table:buliding-notations}
\end{center}
\end{table}
The dynamic system model is given as
\begin{align*}
&C_2 \dot {\bf T}_{{\rm a},t}= \frac{{\bf T}_{{\rm o},t} - {\bf
T}_{{\rm a},t}}{R_2}+\frac{{\bf T}_{{\rm w},t} - {\bf T}_{{\rm a},t}}{R_1}\\
&+(1-a){\bf q}_{{\rm solar},t} + {\bf q}_{{\rm HVAC},t} + {\bf q}_{{\rm internal},t},\\
&C_1 \dot {\bf T}_{{\rm w},t} = \frac{{\bf T}_{a,t} - {\bf
T}_{w,t}}{R_1}+ \frac{{\bf T}_{{\rm o},t} - {\bf T}_{{\rm
w},t}}{R_3} + a{\bf q}_{{\rm solar},t}.
\end{align*}

A discrete time representation can be obtained by using the Euler
discretization with a sampling time of $\Delta t$
\begin{align}
&{\bf T}_{{\rm a},k+1} - {\bf T}_{{\rm a},k} = \frac{\Delta t}{C_2
R_2 }({\bf T}_{{\rm o},k} - {\bf T}_{{\rm a},k})\nonumber\\
&+ \frac{\Delta t}{C_2 R_1}({\bf T}_{{\rm w},k} - {\bf T}_{{\rm
a},k}) + \frac{\Delta t(1-a)}{C_2} {\bf q}_{{\rm solar},k}\nonumber\\
& +\frac{\Delta t}{C_2} {\bf q}_{{\rm HVAC},k} + \frac{\Delta
t}{C_2}{\bf q}_{{\rm internal},k},\nonumber\\
&{\bf T}_{{\rm w},k+1} - {\bf T}_{{\rm w},k} = \frac{\Delta t}{C_1
R_1 }({\bf T}_{{\rm a},k} - {\bf T}_{{\rm w},k})\nonumber\\
& + \frac{\Delta t}{C_1 R_3}({\bf T}_{{\rm o},k} - {\bf T}_{{\rm
w},k})+\frac{\Delta ta}{C_1}{\bf q}_{{\rm
solar},k},\label{eq:building-model}
\end{align}
where $k \geq 0$ is the discrete time step. Here, we consider
$\Delta t = 10 {\rm min}$ sampling time, which is a finer
time-scale than the time step $\Delta t = 30 {\rm min}$ which is
usually used in the building research literature in order to
consider changes of occupant's thermal comfort.

\subsection{Occupant and Weather Models}\label{subsec:occupant-model}
\begin{figure}[h]
\centering\epsfig{figure=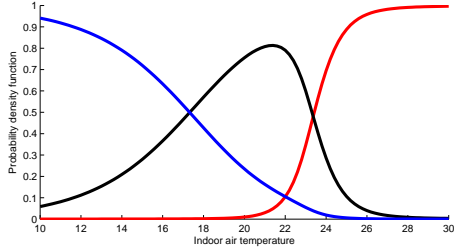,height =3.5cm,width=7cm}
\caption{The probability density function $p(1;{\bf T}_{\rm a})$
(blue), $p(2;{\bf T}_{\rm a})$ (black),  $p(3;{\bf T}_{\rm a})$
(red) for different $T_{\rm a}$}\label{fig:probability-plot}
\end{figure}

A stochastic process $({\bf z}_k)_{k=0}^\infty$ with the
state-space $\{1,2,3\}$ represents the occupant's feelings of
cold, comfort, and hot, respectively. Its probability mass
function $p(i;{\bf T}_{\rm a})$, $i \in \{1,2,3\}$, depending on
the air temperature, ${\bf T}_{\rm a}$, is obtained by the
Bayesian modelling approach~\cite{lee2017bayesian} and is depicted
in~\cref{fig:probability-plot} for different ${\bf T}_{\rm a}$.
We consider the stochastic function ${\bf M}_k:\{0,1,\ldots,143\}
\to\{0,1\}$ to represent the occupancy at time step $k\in \{0,1,
\ldots,143\}$. The occupant arrives at the room at time ${\bf
w}_1$ which is a random variable uniformly distributed within
$\{48,\ldots,54\}$ (between 8am and 9am) and leaves the room at
time ${\bf w}_2$ which is a random variable uniformly distributed
within $\{96,\ldots,114\}$ (between 4pm and 7pm). ${\bf M}_k=1$ if
the room is occupied at time $k$ and ${\bf M}_k=0$ otherwise.

The 24 hours real weather histories $({\bf T}_{{\rm o},k},{\bf
q}_{{\rm solar},k})_{k=0}^{143}$ over 31 days are collected during
July, 2017, in West Lafayette, Indiana, USA, and are used to
approximate them into a Markov chain. In particular, the outdoor
temperature within the range $[10,40]$ is discretized into 6
points, $12.5$, $17.5$,$22.5$, $27.5$, $32.5$, and $37.5$ and
augmented with the periodic time step $k \in\{0,1,\ldots,143\}$,
which is periodically initialized to $k=0$ after the last time
step $k= 143$, to form $6 \times 144$ the state-space. We
constructed a Markov chain with this state space. Similarly, the
solar radiation within the range $[0,900]$ is discretized into 9
points, $50$, $150,\ldots,850$ and augmented with time $k$ to form
a state-space with size $9 \times 144$. A Markov chain with this
state-space was constructed in a similar way.

\subsection{Overall State-Space Model}
From the discretized model~\eqref{eq:building-model}, we obtain a
state-space model ${\bf x}_{k+1}=A {\bf x}_k + B{\bf u}_k + D{\bf
x}_k$ with ${\bf u}_k = {\bf q}_{{\rm HVAC},k}$,
\begin{align}
&{\bf x}_k = \begin{bmatrix}
      {\bf T}_{{\rm a},k} \\
   {\bf T}_{{\rm w},k}\\
\end{bmatrix},\,{\bf w}_k = \begin{bmatrix}
   {\bf q}_{{\rm solar},k} \\
   {\bf q}_{{\rm internal},k} \\
   {\bf T}_{{\rm o},k}\\
\end{bmatrix},\label{eq:x-w}
\end{align}
and
\begin{align*}
&A = \begin{bmatrix}
   1 - \frac{\Delta t}{C_2 R_2} - \frac{\Delta t}{C_2 R_1} & \frac{\Delta t}{C_2 R_1}\\
   \frac{\Delta t}{C_1 R_1} & 1 - \frac{\Delta t}{C_1 R_3} - \frac{\Delta t}{C_1 R_1}\\
\end{bmatrix},\\
&B = \begin{bmatrix}
   \frac{\Delta t}{C_2}  \\
   0  \\
\end{bmatrix},\quad D = \begin{bmatrix}
   \frac{\Delta t(1 - a)}{C_2} & \frac{\Delta t}{C_2} & \frac{\Delta t}{C_2 R_2} \\
   \frac{\Delta ta}{C_1} & 0 & \frac{\Delta t}{C_1 R_3}\\
\end{bmatrix}.
\end{align*}
The internal heat is ${\bf q}_{{\rm internal},k}= 75 + 70{\bf
M}_k$ ($W$), where the first term, $75$, is internal heat due to
electronic appliances, and the second term, $70{\bf M}_k$,
indicates the heat produced by the occupant. The complete discrete
state including the weather conditions is
\begin{align*}
&{\bf s}_k = \begin{bmatrix}
  {\bf x}_k^T & {\bf z}_k & (k\,{\rm mod}\,144) & {\bf T}_{{\rm o},k} & {\bf q}_{{\rm solar},k} & {\bf M}_k \\
\end{bmatrix}^T,
\end{align*}
where $(k\, {\rm mod}\,144)$ is the remainder of $k$ divided by
$144$ and represents the 24 hours periodic time steps. We assume
that the occupant's comfort level ${\bf z}_k$, wall temperature
${\bf T}_{{\rm w},k}$, outdoor temperature ${\bf T}_{{\rm o},k}$,
and solar radiation ${\bf T}_{{\rm solar},k}$ cannot be observed.
Therefore, the observation is
\begin{align}
&{\bf o}_k = \begin{bmatrix}
   {\bf T}_{{\rm a},k} & (k\, {\rm mod}\,144) & {\bf M}_k \\
\end{bmatrix}^T.\label{eq:building-observation}
\end{align}

\subsection{Reward Design}

In general, the reward function for RL-based designs is
hand-engineered with trial and errors to demonstrate reasonable
control performances. Since the performance of the learned policy
depends considerably on the reward, there exist approaches to
design rewards, for instance, the inverse
RL~\cite{ng2000algorithms}. However, these approaches will not be
considered here. We first introduce a hand-crafted cost function
${\bf c}({\bf s}_k,{\bf u}_k)$, and then set ${\bf r}({\bf
s}_k,{\bf u}_k)= -{\bf c}({\bf s}_k,{\bf u}_k)$ to minimize the
long term discounted costs. In particular, the cost function is
set to be
\begin{align*}
&{\bf c}({\bf s}_k,{\bf u}_k): = \begin{cases}
   0.001{\bf u}_k^2 \quad {\rm if}\,\,\,{\bf M}_k = 0 \\
   0.00001{\bf u}_k^2  + {\bf c}_{1,k} + {\bf c}_{2,k} \quad {\rm if}\,\,\,{\bf M}_k = 1  \\
\end{cases}
\end{align*}
where ${\bf c}_{1,k}$ is the penalty term related to the state
constraint ${\bf T}_{a,k}\in [20,30]$:
\begin{align*}
&{\bf c}_{1,k} = \begin{cases}
 0\quad {\rm if}\,\,{\bf T}_{a,k}\, \in [20,30] \\
 200\quad {\rm otherwise} \\
 \end{cases}
\end{align*}
and ${\bf c}_{3,k}$ is the penalty term related to the comfort of
the occupant:
\begin{align*}
&{\bf c}_{2,k} = \begin{cases}
 0\quad {\rm if}\,\,{\bf z}_k = 2 \\
 100\quad {\rm if}\,\,{\bf z}_k \in \{ 1,3\}  \\
 \end{cases}
\end{align*}

\subsection{Simulation}

\begin{figure*}[h]
\centering\epsfig{figure=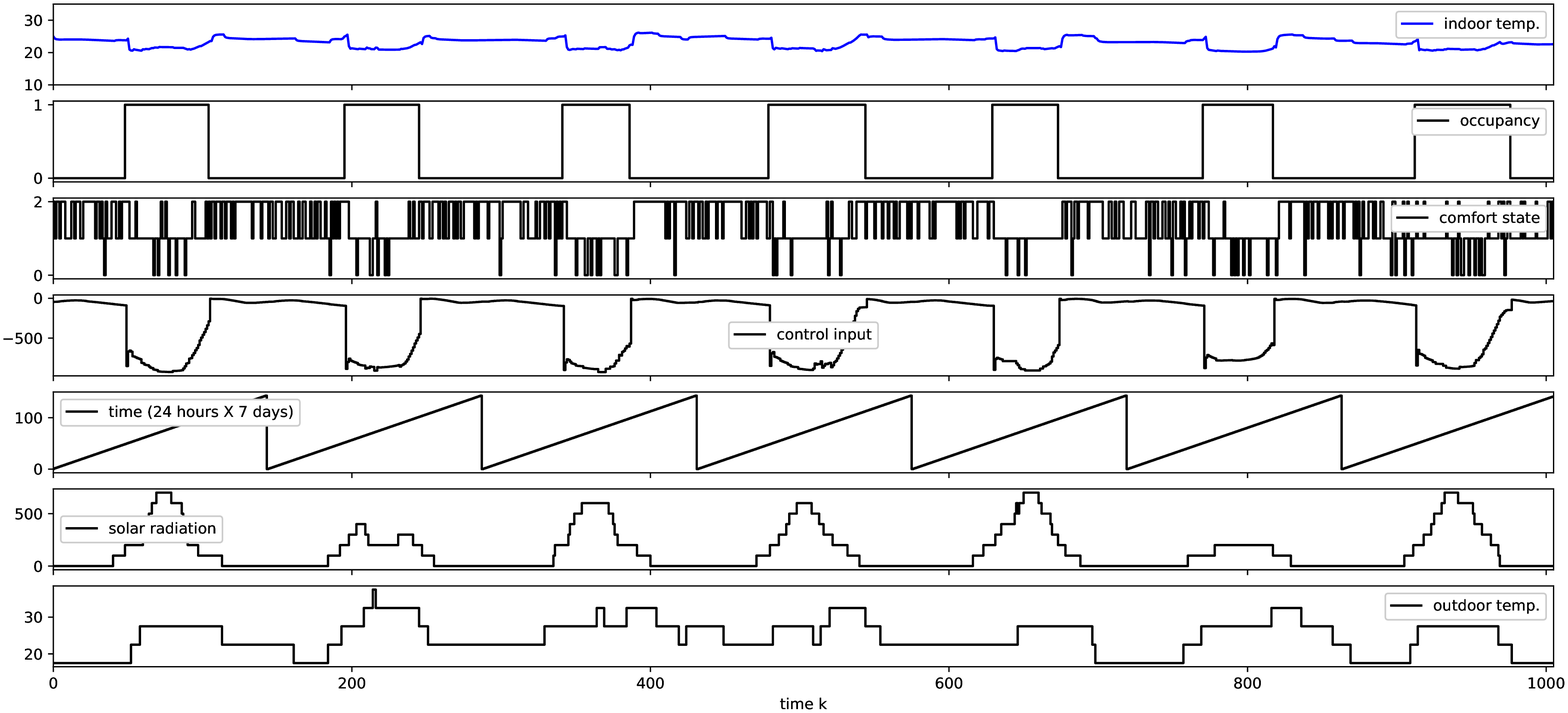,width=\linewidth,height=7cm}
\caption{One week simulation results with a policy learned by
DDPG. The indoor air temperature (blue solid line), occupancy,
control input, time steps which reset every 24 hours, solar
radiation, and outdoor air temperature from the top to bottom
figures.}\label{fig:one-week-simulation}
\end{figure*}

\begin{figure*}[h]
\centering\epsfig{figure=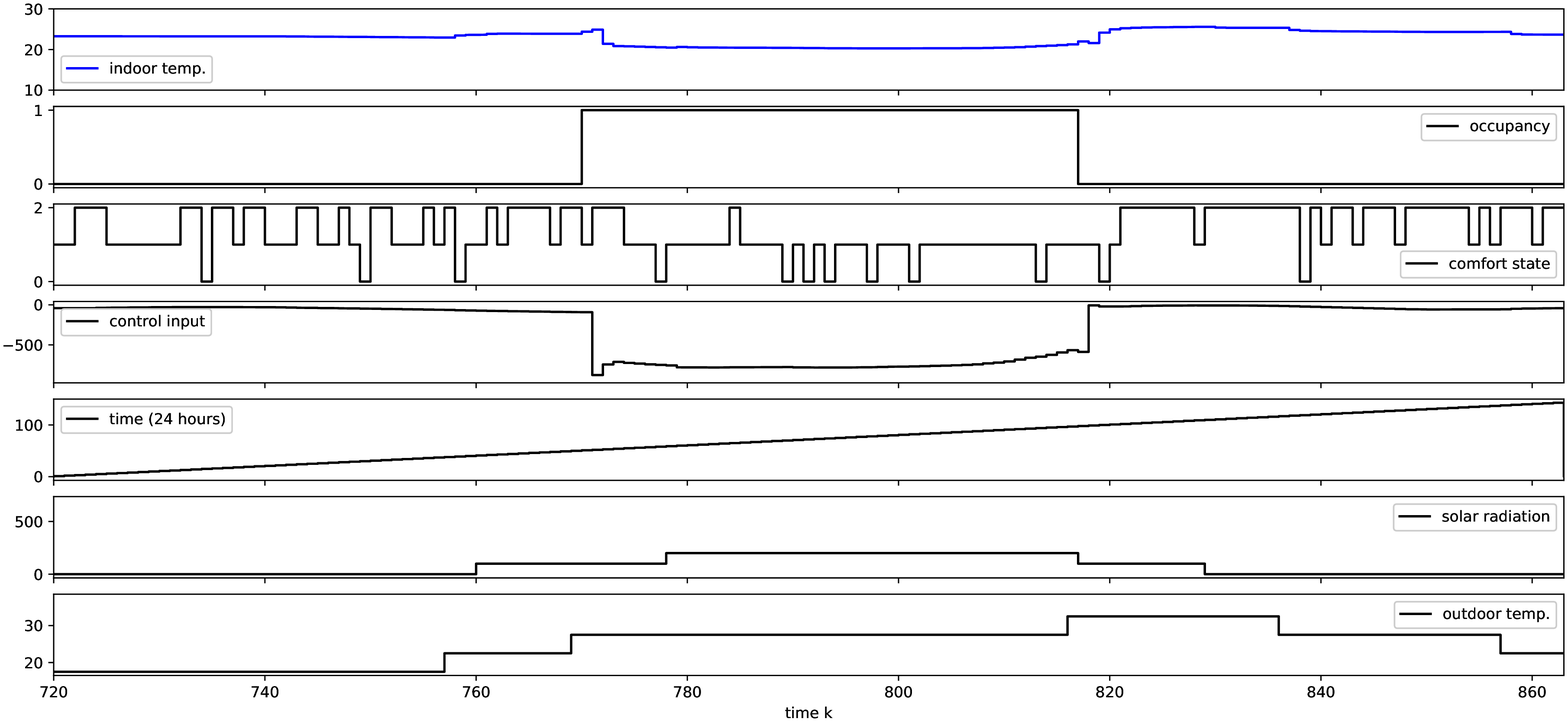,width=\linewidth,height=7cm}
\caption{The fifth day results
of~\cref{fig:one-week-simulation}.}\label{fig:one-day-simulation}
\end{figure*}

\begin{figure*}[h]
\centering\epsfig{figure=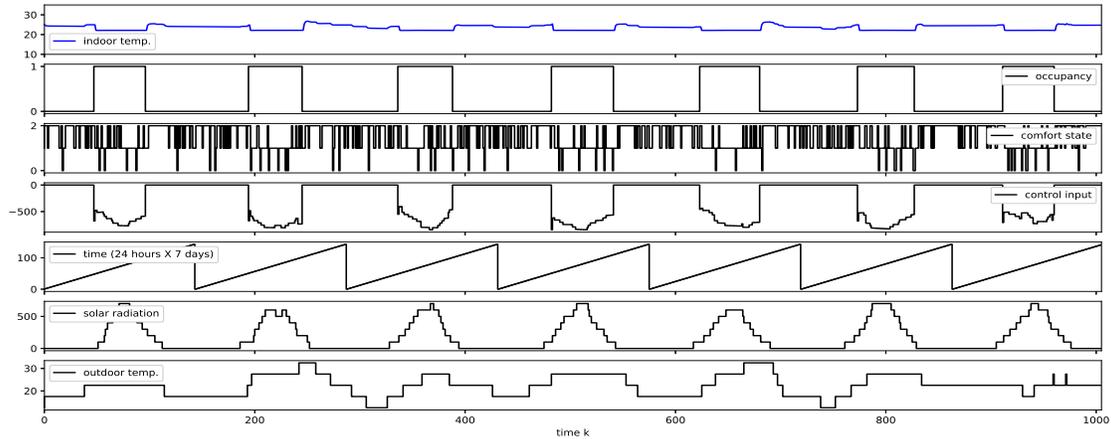,width=\linewidth,height=7cm}
\caption{One week simulation results with the greedy
policy~\eqref{eq:greedy-control}. The indoor air temperature (blue
solid line), occupancy, control input, time steps which reset
every 24 hours, solar radiation, and outdoor air temperature from
the top to bottom figures.}\label{fig:one-week-simulation2}
\end{figure*}

\begin{figure*}[h]
\centering\epsfig{figure=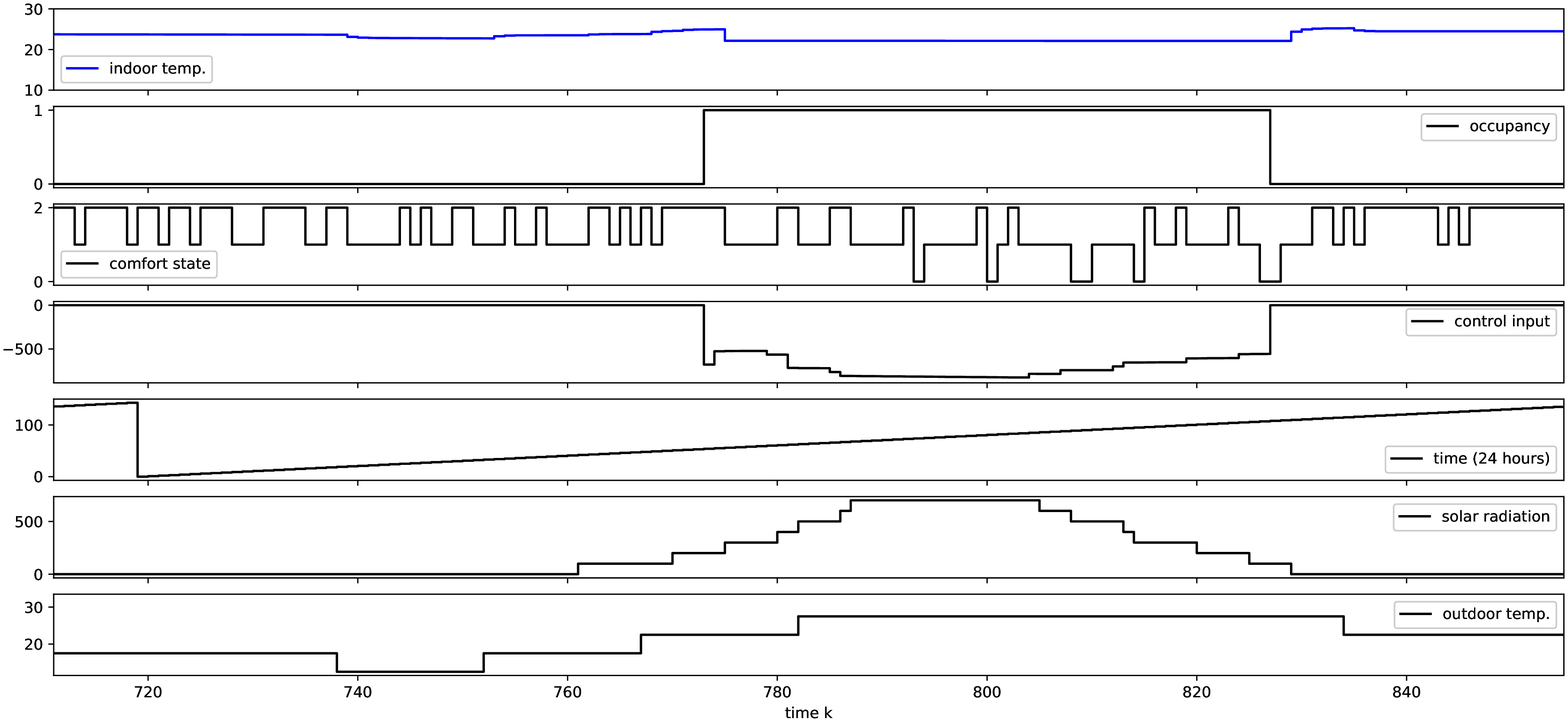,width=\linewidth,height=7cm}
\caption{The fifth day results
of~\cref{fig:one-week-simulation2}.}\label{fig:one-day-simulation2}
\end{figure*}

For simulation studies, we implement the DDPG using Python. This
work uses neural networks with four hidden layers of width
$(1024,1024,512,512)$ for both actor and critic networks. We
implemented~\cref{algo:DDPG} with approximately two hours of
training. Simulation results of a one week operation under the
learned policy are given in~\cref{fig:one-week-simulation}. The
fifth day's results in~\cref{fig:one-week-simulation} are plotted
in~\cref{fig:one-day-simulation}, which shows that without
observations of the outdoor temperature and solar radiation, the
trained policy infers the decrease of the the outdoor
temperature/solar radiation, and starts to decrease the input
power. For a comparative analysis, a greedy policy is considered,
which minimizes a cost function based on the one-step-ahead
prediction of the system trajectory. In particular, the greedy
policy ${\bf u}_k=\mu({\bf x}_k,{\bf w}_k,{\bf M}_k)$ is defined
as
\begin{align}
&\mu ({\bf x}_k ,{\bf w}_k ,{\bf M}_k): = \begin{cases}
   {\bf u}_k^* \quad {\rm if}\,\,{\bf M}_k  = 1\\
   0\quad {\rm otherwise}  \\
\end{cases}
\label{eq:greedy-control}
\end{align}
where
\begin{align*}
&{\bf u}_k^*:= \argmin_{|u| \le 1000} \{ \gamma ([{\bf x}_{k+1}
]_1 -22)^2  + 10^{-5} u^2 \},
\end{align*}
$[{\bf x}_{k+1}]_1$ is the first element of ${\bf x}_{k+1}$, ${\bf
x}_{k+1}= A{\bf x}_k + B{\bf u}_k + D{\bf w}_k$, and the number 22
is the temperature that maximizes the comfort probability, i.e.,
$22 \cong \argmax_{T_a } p(2;T_a)$. The policy tries to enforce
the current indoor air temperature to be 22$C^\circ$ while
minimizing the current control input energy. One week simulation
results under the greedy control policy are given
in~\cref{fig:one-week-simulation2}. The fifth day's results are
plotted in~\cref{fig:one-day-simulation2}.

\begin{figure}[h]
\centering\epsfig{figure=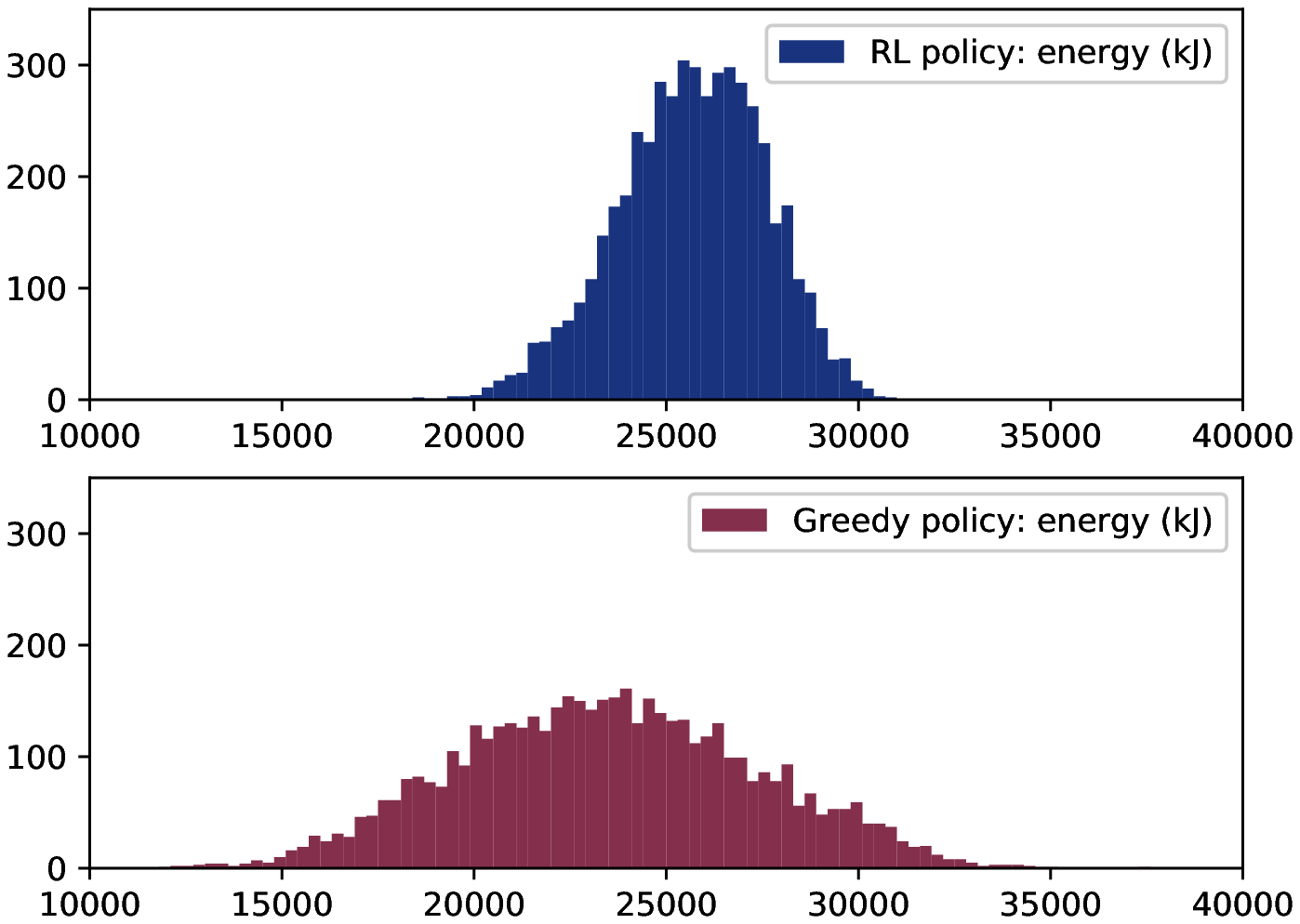,width=8cm,height=6cm}
\caption{Histograms of the total input energy (kJ) during a single
day for the RL policy (top figure) and the greedy policy (bottom
figure).}\label{fig:energy-histogram}
\end{figure}

\begin{figure}[h]
\centering\epsfig{figure=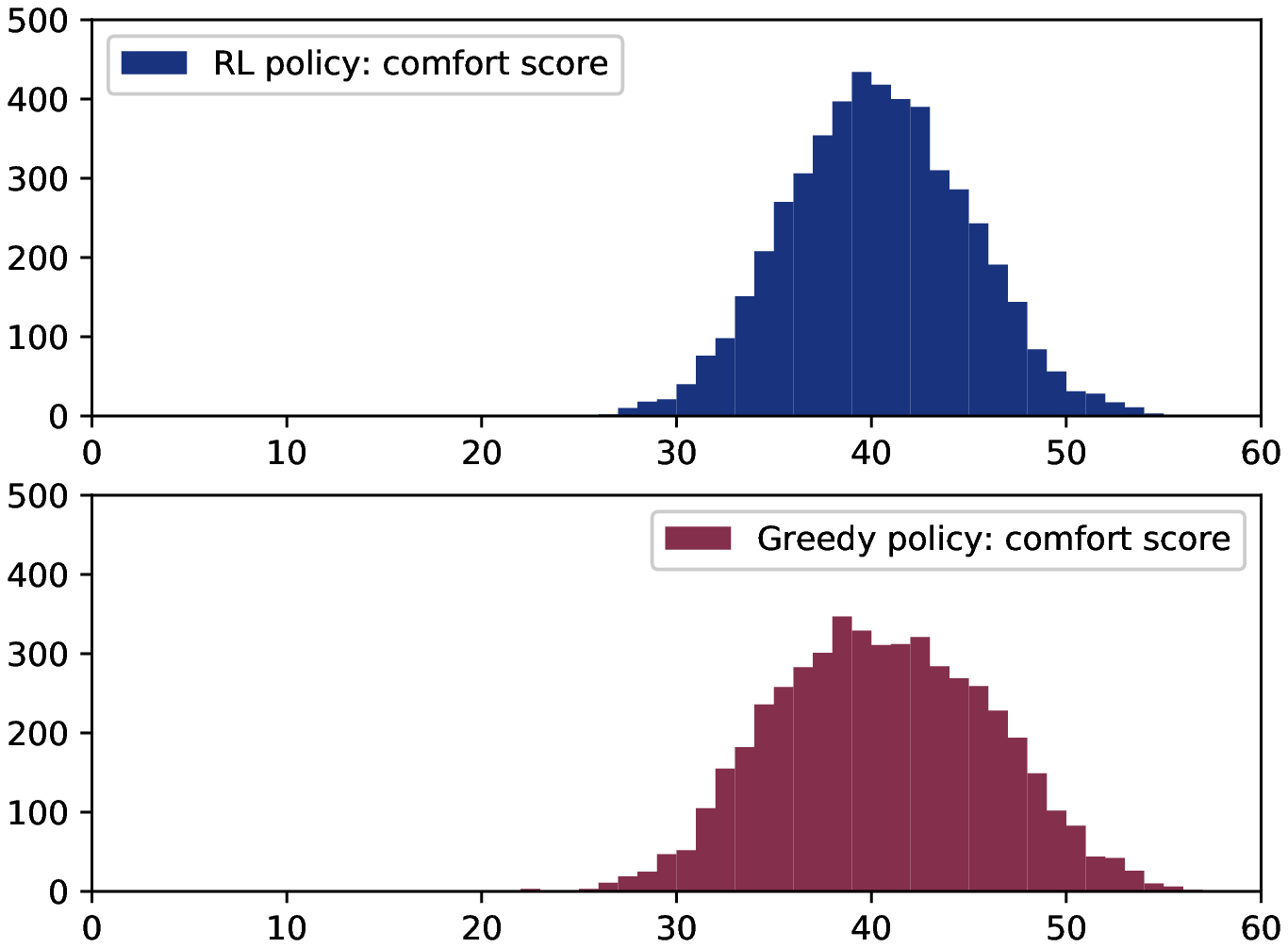,width=8cm,height=6cm}
\caption{Histograms of the total comfort scores during a single
day for the RL policy (top figure) and the greedy policy (bottom
figure).}\label{fig:comfort-histogram}
\end{figure}

Histograms of the total input energy (kJ) during a single day are
depicted in~\cref{fig:energy-histogram} for the RL policy (top
figure) and the greedy policy (bottom figure) with 5000 samples.
In addition, histograms of the total comfort scores during a
single day are illustrated in~\cref{fig:comfort-histogram} for the
RL policy (top figure) and the greedy policy (bottom figure).
During a single day, the comfort score is computed by counting the
total number of comfort feelings, i.e., ${\bf z}_k = 2$, when the
room is occupied. From the results, one concludes that both
policies demonstrate similar performances. However, the policy
learned by DDPG only uses a direct
observation~\eqref{eq:building-observation}, while the greedy
policy~\eqref{eq:greedy-control} requires the full knowledge of
the system state at every time steps.

\section*{Conclusion}
The building climate control problem with occupant interactions is
an active topic. This paper has presented an application of DDPG
approach to the building problem and assessed its performance.
Comparative simulation results have suggested that the policy
learned by DDPG outperforms other methods. The results were based
on the simulation-based off-line learning, which is possible only
when an overall model of the building system is known. Future
research will focus on a real-world implementation of the RL with
occupant interactions. In this case, the real-time RL requires a
real-time user feedback interface with a voting system to learn
the policy based on the occupant's satisfaction.

\bibliographystyle{IEEEtran}
\bibliography{reference}

\end{document}